\documentclass{article}

\usepackage[final,nonatbib]{neurips_2021}

\usepackage[utf8]{inputenc} % allow utf-8 input
\usepackage[T1]{fontenc}    % use 8-bit T1 fonts
\usepackage{hyperref}       % hyperlinks
\usepackage{url}            % simple URL typesetting
\usepackage{booktabs}       % professional-quality tables
\usepackage{amsfonts}       % blackboard math symbols
\usepackage{nicefrac}       % compact symbols for 1/2, etc.
\usepackage{microtype}      % microtypography
\usepackage[table,dvipsnames]{xcolor}

\usepackage{graphicx} %package to manage image
\usepackage{amssymb}
\usepackage{threeparttable}
\usepackage{multicol}
\usepackage{multirow}
\usepackage{colortbl}
\usepackage{bm}
\usepackage{enumitem}
\usepackage{algorithm}% http://ctan.org/pkg/algorithms
\usepackage{algpseudocode}% http://ctan.org/pkg/algorithmicx 
\usepackage{blindtext}
\usepackage{amsmath}
\usepackage{wrapfig}
\usepackage{capt-of}
\usepackage{tabularx}
\usepackage[export]{adjustbox}

\newcommand{\etal}{\textit{et al}. }

\DeclareMathOperator*{\argmin}{arg\,min}

\definecolor{mygreen}{rgb}{0.2,0.5,0.2}

\title{Mosaicking to Distill: \\ Knowledge Distillation from Out-of-Domain Data}
\author{
Gongfan Fang$^{1,4}$, Yifan Bao$^1$, Jie Song$^1$, Xinchao Wang$^2$, Donglin Xie$^1$ \\ \bf Chengchao Shen$^3$, Mingli Song$^{1}$\thanks{Corresponding author.} \\
$^1$Zhejiang University, $^2$National University of Singapore, $^3$Central South University\\
$^4$Alibaba-Zhejiang University Joint Institute of Frontier Technologies\\
\texttt{\{fgf,yifanbao,sjie,donglinxie,brooksong\}@zju.edu.cn} \\
\texttt{xinchao@nus.edu.sg}, \texttt{scc.cs@csu.edu.cn}\\
}

\begin{document}

\maketitle

\begin{abstract}
Knowledge distillation~(KD) aims to 
craft a compact student model
that imitates the behavior 
of a pre-trained teacher in a target domain.
Prior KD approaches, despite their gratifying results,
have largely relied on the premise that 
\emph{in-domain} data is available to carry
out the knowledge transfer.
Such an assumption, unfortunately, 
in many cases violates the practical setting,
since the original training data or even the data
domain is often unreachable due to privacy or
copyright reasons. In this paper, we attempt to
tackle an ambitious  task, termed as 
\emph{out-of-domain} knowledge distillation~(OOD-KD),
which allows us
to conduct KD using only OOD data
that can be readily obtained at a very low cost.
Admittedly,  OOD-KD is by nature
a highly challenging task 
due to the agnostic domain gap.
To this end, 
we introduce a handy yet 
surprisingly efficacious approach, 
dubbed as~\textit{MosaicKD}.
The key insight behind MosaicKD lies in that,
samples from various domains share
common local patterns,
even though their global semantic may 
vary significantly;
these shared local patterns,
in turn, can be re-assembled
analogous to mosaic tiling,
to approximate the in-domain data
and to further alleviating
the domain discrepancy.
In MosaicKD, this 
is achieved through a four-player min-max game,
in which a generator, a discriminator,
a student network, 
are  collectively trained in an adversarial manner,
partially under the guidance of a pre-trained teacher.
We validate MosaicKD over 
{classification and semantic segmentation tasks}
across various benchmarks,
and demonstrate that
it yields results much superior to 
the state-of-the-art counterparts on OOD data. 
Our code is available at \url{https://github.com/zju-vipa/MosaicKD}.
\end{abstract}
    
\section{Introduction}
    % Background
    Knowledge distillation~(KD) has emerged as a
    popular paradigm for model compression and knowledge transfer,
    attracting attention from various research 
    communities~\cite{hinton2015distilling,romero2014fitnets,tian2019contrastive,gou2020knowledge}.
    The goal of KD is to train a lightweight model, known as the student,
    by imitating a pre-trained but more cumbersome model, 
    known as the teacher, 
    so that the student masters the expertise of the teacher.
    In recent years, KD has demonstrated encouraging results 
    over various machine learning applications,
    including but not limited to computer vision~\cite{chen2017learning,li2020gan}, 
    data mining~\cite{arora2019knowledge}, and natural language processing~\cite{sun2019patient,jiao2019tinybert} %~\xw{\cite{XXX}}.
    
    Nevertheless, the conventional setup 
    for KD has largely relied on the premise that,
    data from at least the same domain, 
    if not the original training data,
    is available to train the student. 
    This seemingly-mind assumption,
    paradoxically, imposes a major constraint 
    for conventional KD approaches:
    in many cases, the training data and even their domain
    for a pre-trained network are agnostic,
    due to for example confidential or copyright reasons. 
    Hence, the in-domain prerequisite significantly
    limits the applicable scenarios of KD,
    and precludes taking advantage of
    the sheer number of publicly-available pre-trained models,
    many of which with unknown training domain~\cite{kolesnikov2019big,radford2019language},
    to carry out massive knowledge transfer.
    
    In this paper, we aim at the ambitious goal of conducting KD using
    only \emph{out-of-domain}~(OOD) data, which,
    in turn, enables us to greatly relax the conventional prerequisite
    and thereby largely strengthens applicability of KD.
    Unarguably, OOD-KD is by nature a highly challenging task, since
    the domain discrepancy will inevitably impose a major obstacle
    towards the proper functioning of the pre-trained teacher.
    In fact, if we are to conduct naive KD on the raw OOD data,
    the resulting student model, {as will be demonstrated in our experiments},
    fails to provide any performance guarantee on the target domain.
    This  phenomenon signifies the limited generalization capability learned from OOD data,
    which is unsurprising.

\begin{wrapfigure}{r}{0.6\textwidth}
      \begin{center}\vspace{-5mm}
          \includegraphics[width=\linewidth]{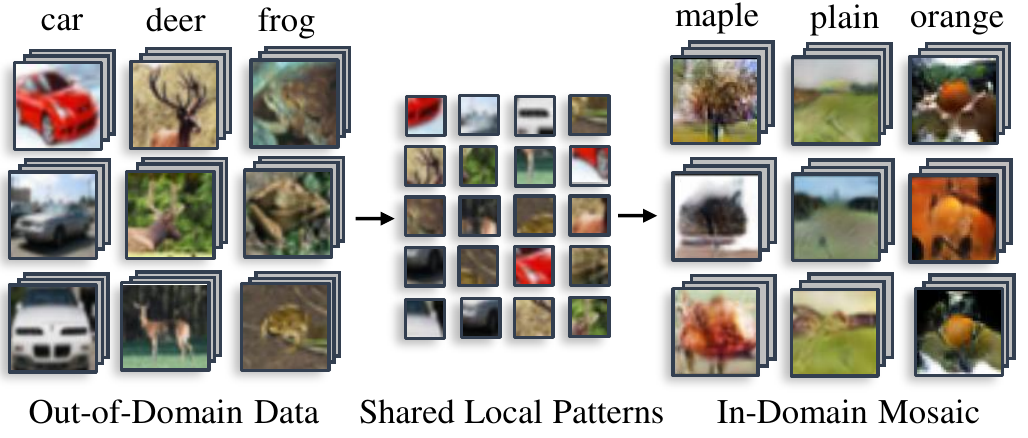}
      \end{center}\vspace{-2mm}
          \caption{Natural images share common local patterns.
          In MosaicKD, 
          these local patterns are first dissembled from OOD data
          and then assembled to synthesize in-domain data, making
          OOD-KD feasible.
          } \label{fig:intro}
          \vspace{-2mm}
    \end{wrapfigure}

    To this end, we propose a novel assembling-by-dismantling 
    approach, termed 
    as MosaicKD,
    that allows us to 
    take advantage of  OOD data
    to conduct KD.
    Our motivation  stems from the fact that, 
    even though data from different domains 
    exhibit divergent global distributions,
    their local distributions, such as patches 
    in images, may however resemble each other. 
    %In Fig.~\xw{XXX}, for example, \xw{XXX}.
    This observation further inspires us to leverage 
    the local patterns, shared by the OOD  and target-domain data,
    to resolve the domain shift problem in OOD-KD.
    As such, the core idea of MosaicKD is
    to synthesize in-domain data, of which  the local patterns 
    imitate those from real-world OOD data, 
    while the global distribution,
    assembled from local ones,    
    is expected to fool the pre-trained teacher. 
    As shown in Figure \ref{fig:intro}, the shared local patterns are extracted from
    OOD data and re-assembled into in-domain data. 
    %two diverged domains can be bridged by their shared local patterns. 
    Intuitively, this process
    is analogous to mosaic tiling,
    where tesserae
    are utilized to compose the whole art piece. 
    
    % Method
    Specifically, in MosaicKD, we frame OOD-KD problem as a
    novel four-player min-max game involving
    a generator, a discriminator, a student, and a teacher,
    among which the former three are to be learned while
    the last one is pre-trained and hence fixed.
    The generator, as those in prior GANs, 
    takes as input a random noise vector
    and learns to mosaic synthetic in-domain samples
    with locally-authentic and globally-legitimate
    distributions, under the supervisions back-propagated from
    the other three players.
    The discriminator, on the other hand,
    learns to distinguish local patches
    extracted from the real-world OOD data 
    and from the synthetic samples.  
    The entire synthetic images are fed to 
    both the pre-trained teacher and the to-be-trained
    student, based on which the teacher 
    provides category knowledge for data synthesis
    %\xwc{pr the global semantic of synthetic data}
    %\xwc{``align'' is very fuzzy. Change to another word}
    and the student
    mimics the behavior of the teacher
    so as to carry out KD.
    The four players collaboratively
    reinforce one another
    in an adversarial fashion,
    and collectively
    accomplish the student training.

    In short, our contribution is 
    the first dedicated attempt towards
    OOD-KD,  a highly practical 
    yet largely overlooked problem, 
    achieved through a novel scheme
    that mosaics in-domain data.
    The synthetic samples,
    generated via a four-player min-max game,
    enjoy realistic local structures
    and sensible
    global semantics,
    laying the ground for 
    a dependable knowledge distillation
    from the pre-trained teacher.
    We conduct experiments over 
    classification and semantic segmentation
    tasks across various benchmarks,
    and demonstrate that MosaicKD
    yields truly encouraging results
    much superior to those derived
    by its state-of-the-art competitors
    on OOD data.
    %\xw{sometimes even approaching
    %the KD results on in-domain data.}

\section{Related Work}
  
    \textbf{Knowledge distillation.} Knowledge Distillation aims to craft a lightweight student model from cumbersome teachers~\cite{hinton2015distilling}, by either transferring network outputs~\cite{bucilua2006model,hinton2015distilling} or intermediate representations~\cite{romero2014fitnets,park2019relational}. In the literature, knowledge distillation and its variants largely rely on the premise that original training data is available during distillation, which is vulnerable in real-world applications due to privacy or copyright reasons~\cite{micaelli2019zero,yang2020distillating,shen2021progressive,yang2020factorizable,shen2019customizing,jing2021amalgamating,shen_AAAI_2019}. Recently, data-free knowledge distillation~\cite{lopes2017data,micaelli2019zero,fang2019data,yin2019dreaming,Ye_2020_CVPR} has attracted attention from various research communities, which trains student model only with synthetic data. However, due to the difficulty in data synthesis without real-world samples, data-free KD usually leads to a degraded student on complicated tasks~\cite{yin2019dreaming,chawla2021data}. Another solution to relax the conventional prerequisite on training data is to use some OOD samples. Unfortunately, it is found that naive knowledge distillation on OOD usually fails to learn a comparable student model from teachers~\cite{menghani2019learning,kulkarni2017knowledge}. 
    
    \textbf{Domain adaptation and generalization.} Most learning algorithms strongly rely on the premise that the source data for training and the target data for testing are independent and identically distributed~\cite{vapnik1992principles}, ignoring the OOD problem that is frequently encountered in real-world applications. In the literature, the OOD problem is usually addressed by domain generalization (DG) or adaptation (DA)~\cite{blanchard2011generalizing,zhou2021domain}. Adaptation is a popular technique for aligning the source and target domain~\cite{saito2018maximum,ganin2015unsupervised,tzeng2017adversarial}, which typically requires the target domain to be accessible during training. In recent years, Domain adaptation has been extended to open-set settings where the label space of training and testing data are different~\cite{panareda2017open}. In comparison, domain generalization is similar to domain adaptation, but does not requires the information from the target domain~\cite{blanchard2011generalizing}. Domain generalization trains a model on the source domain only once and directly applied the model on the target domain~\cite{li2018domain,ghifary2015domain,volpi2018generalizing,NEURIPS2019_8a50bae2}. Despite the success of DA and DG in supervised learning, OOD problem is still under-studied in the context of knowledge distillation.%, where domain information can be extracted from an already trained teacher model.   
    
    \textbf{Generative adversarial networks (GAN).} Generative adversarial network is initially introduced by Goodfellow \etal for image generation~\cite{goodfellow2014generative}, where a generator is trained to fool a discriminator in an adversarial minimax game. In recent years, several works have been proposed to improve the performance of GANs, from the perspective of image quality~\cite{karras2017progressive,karras2019style}, data diversity~\cite{mao2019mode,yang2019diversity}, and training stability~\cite{arjovsky2017wasserstein}. Besides, Some explorations have been taken to make GAN training more efficient with limited data via transfer learning~\cite{wang2018transferring} or augmentation~\cite{zhao2020differentiable}. In this work, we study data synthesis in OOD settings, where the original training data is unavailable and thus conventional GAN technique can not be diretly deployed. 

\section{Out-of-Domain Knowledge Distillation}

\newcommand{\distX}[1][]{\mathcal{X}}
\newcommand{\distY}[1][]{\mathcal{Y}}

Without loss of generality, we study OOD problem in the context of image classification task. The domain underlying a dataset is defined as a triplet $\mathcal{D}=\{\distX, \distY, P_{\distX\times\distY}\}$, consisting of input space $\distX \subset \mathbb{R}^{c\times h\times w}$, label space $\distY=\{1, 2, ..., K\}$ and joint distribution $P_{\distX\times\distY}$ over $\distX\times\distY$. Given a teacher model $T(x; \theta_t)$  optimized on target domain $\mathcal{D}$, vanilla KD trains a lightweight student model to imitate teacher's behaviour, by directly minimizing the empirical risk on the original domain:
\begin{equation}
    \theta_s^* = {\argmin}_{\theta_s} \; \mathbb{E}_{ (x, y) \sim P_{\distX\times\distY}} \left[  \ell_{\text{KL}}(T(x;\theta_t) \| S(x; \theta_s)) + \ell_{\text{CE}}(S(x;\theta_s), y ) \right], \label{eqn:vanilla_kd}
\end{equation}
where $\ell_{\text{KL}}$ and $\ell_{\text{CE}}$ refers to the KL divergence and the cross entropy loss. However, when the original training domain $\mathcal{D}$ is unavailable and some alternative data from another domain $\mathcal{D}^{\prime}=\{\distX^{\prime}, \distY^{\prime}, P_{\distX\times\distY}\}$ is used for training, Equation \ref{eqn:vanilla_kd} may be problematic if the domain gap is huge. In this work, we focus on the out-of-domain problem in knowledge distillation, described as follows: 

\textbf{Problem Definition (OOD-KD).}  Given a teacher model $T(x; \theta_t)$ obtained from training domain $\mathcal{D}=\{\distX, \distY, P_{\distX\times\distY}\}$, the goal of OOD-KD is to craft a student model $S(x; \theta_s)$ only leveraging out-of-domain data from  $\mathcal{D}^{\prime}=\{\distX^{\prime}, \distY^{\prime}, P_{\distX^{\prime}\times\distY^{\prime})}\}$, where $\distX^{\prime} \neq \distX$ and $\distY^{\prime} \neq \distY$.

In OOD-KD, due to the domain divergence between OOD data and original training data, some important patterns may be the missing and the corresponding knowledge on these patterns might not be appropriately transferred from teachers to students. To address the OOD problem, we propose a novel assembling-by-dismantling approach to craft in-domain samples from out-of-domain ones, which effectively alleviates the domain gap between the transfer set and unavailable training set, making KD applicable on out-of-domain data.

\section{Proposed Method}
In the absence of original training data $X$, directly minimizing the risk on an OOD set $X^{\prime}$ would be problematic due to the diverged data domain. In this work, we introduce a generative method for OOD-KD, dubbed as MosaicKD, where a generator $G(z; \theta_g)$ is trained to synthesize a more helpful distribution $P_{G}$ for student learning. Specifically, MosaicKD is developed upon the distributionally robust optimization (DRO) framework which has been widely used to tackle domain shift~\cite{rahimian2019distributionally,NEURIPS2019_8a50bae2,choi2020data}. Given a pre-defined distance metric $d(\cdot, \cdot)$ for distributions, the basic form of DRO framework can be formalized as the following:
\begin{equation}
    \min_S \max_{G}  \;  \{\mathbb{E}_{x \sim P_{G}}\left[ \ell_\text{KL}(T(G(z)) \| S(G(z))) \right] : d(P_G, P_{X^{\prime}}) \leq \epsilon \} \label{eqn:DRO}
\end{equation}
In equation \ref{eqn:DRO}$, \ell_\text{KL}$ denotes the KL divergence for student learning, and $d(P_G, P_{X^{\prime}})$ denotes the distribution distance between the generated samples and OOD data. The hyper-parameter $\epsilon$ specifies the radius of a ball space centered at $P_{X^{\prime}}$. According to this definition, DRO framework aims to find the worst-case distribution from the searching space, which establishes an upper bound for the empirical risk of other distributions covered by the searching space. Ideally, if the target distribution $P_{X}$ of original training data exactly lies in the searching space, its empirical risk can effectively be optimized by the DRO framework. However, we would like to argue that this premise may be problematic in OOD settings, where $\distX \neq \distX^{\prime}$ and $\distY \neq \distY^{\prime}$. Note that if two distributions $P_{X_1}$ and $P_{X_2}$ are close in the input space under certain metric $d(\cdot; \cdot)$, their label space $\distY_1$ and $\distY_2$ should be also similar~\cite{van2020survey}. Based on this, distributions within the small $\epsilon$ ball space, centered at OOD distribution $P_{X^{\prime}}$, are likely to share the same label space, i.e., $\distX \approx \distX^{\prime}$ and $\distY \approx \distY^{\prime}$, which obviously conflicts with the OOD settings. To this end, the target domain of original training data might not be covered by the search space and can not be bounded by the DRO framework. A remedy for this issue is to use a sufficiently large radius $\epsilon$. Unfortunately, it will only lead to intractable searching space, flooded with meaningless distributions. 

\subsection{Mosaicking to Distill}
As mentioned above, the searching space built upon OOD data is insufficient for establishing an reliable upper bound for optimization. To address this problem, MosaicKD introduces a new way to construct the searching space based on local patches.
Our motivation stems from the fact that, patterns of natural images are often organized hierarchically, where high-level patterns are assembled from low-level ones. Although the domain of original training data $X$ and OOD data $X^{\prime}$ are diverged, their local patterns may still resemble each other. For example, the patterns of ``fur'' can be shared by different animal species from varied domains. Note that each images is assembled from local patches, we propose an assembling-by-dismantling strategy to re-organizes shared local patches and synthesize in-domain data for training.

\textbf{Patch Learning.} The first step towards MosaicKD is to extract local patterns from OOD data $X^{\prime}$ and estimate the patch distribution for generation. Given an OOD dataset $X^{\prime}=\{x^{\prime}_1, x^{\prime}_2, ..., x^{\prime}_N; x^{\prime}_i \in \mathbb{R}^{H\times W\times 3}\}$, we obtain patches through $L\times L$ cropping, which produces a patch dataset $C=\{c_1, c_2, ...,c_M; c_i \in \mathbb{R}^{L\times L\times 3}\}$. The patch size $L$ is an important hyper-parameter for MosaicKD. For example, if $L=W=H$, each patch will cover a full image, which contains all high-level features of original images. When we decrease the patch size to $L=1$, then each patch only contains low-level color information. Obviously, small size $L$ can lead to more general patterns than large one, which are more likely to be shared by different domains. Besides, increasing the patch size will introduce more structural information, making the distribution of patches closer to the that of full images. In this work, we model local patch learning as a generative problem, where a generator $G(z; \theta_g)$ is trained to approximate the patch distribution by fooling a discriminator network $D(x; \theta_d)$. Note that our goal is to synthesize full images instead of pieces of patches, we train the generator $G(x; \theta_g)$ to produce images of full resolutions and craft patches on the fly. Let $C(\cdot)$ refers to the cropping operation, the objective of patch learning can be formulated as follows:
\begin{equation}
    \min_G \max_D \mathcal{L}_{local}(G, D) = \mathbb{E}_{x^{\prime} \sim P_{X^{\prime}}} \left[\log D(C(x^{\prime}))\right] + \mathbb{E}_{z\sim P_z} \left[\log(1-D(C(G(z))))\right] \label{eqn:local_gan}
\end{equation}
where $P_{X^{\prime}}$ refers to the distribution of OOD data and $P_z$ refers to the prior distribution of the latent variable $z$. $C(x^{\prime})$ and $C(G(z))$ refers to the cropped patches from OOD data and generated data. The main difference between Eqn.~(\ref{eqn:local_gan}) and the objectives in vanilla GANs~\cite{goodfellow2014generative} lies in the patch-level discrimination, where MosaicKD only regularizes local patterns to be authentic, leaving the global structure unrestricted. As mentioned above, global patterns can be assembled from local ones, MosaicKD assembles these patches to synthesize in-domain data through label space aligning.

\begin{figure*}[t]
  \begin{center}
     \includegraphics[width=0.8\linewidth]{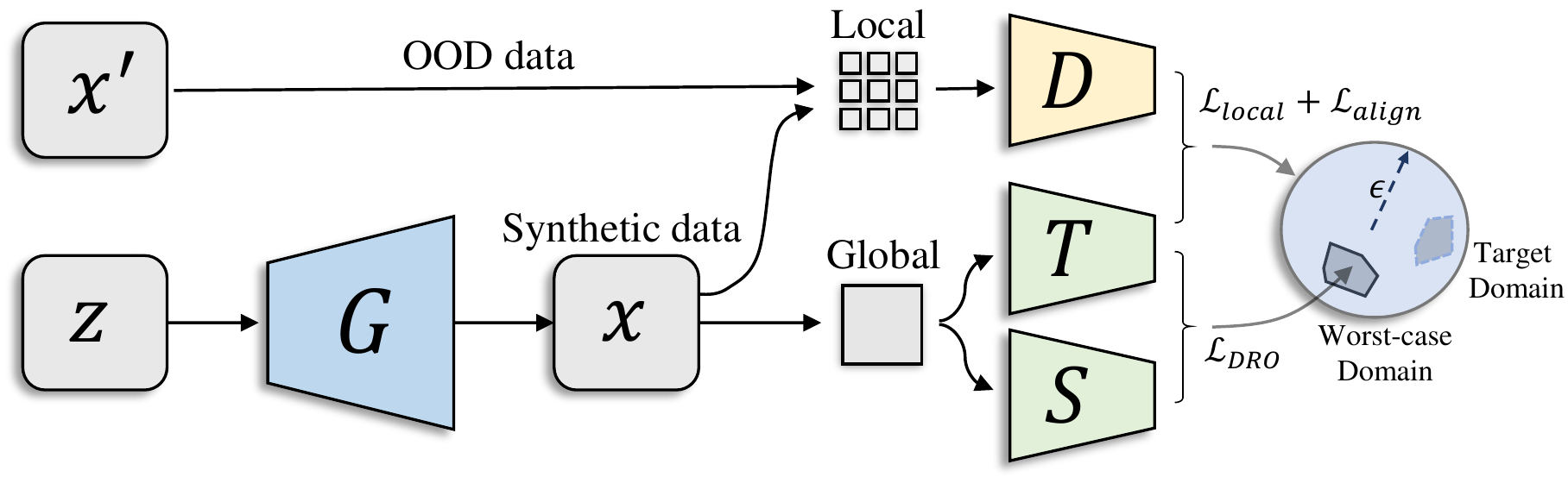}
  \end{center}\vspace{-2mm}
     \caption{The framework of MosaicKD. We leverage the local patterns of OOD data and the category knowledge of teacher to synthesize locally-authentic and globally-legitimate samples for KD.} \label{fig:framework} 
\end{figure*}

\textbf{Label Space Aligning.} As no inter-patch restriction is introduced in Eqn.~(\ref{eqn:local_gan}), the generator may only produce images with meaningless global semantic, although their local patterns are plausible. In this step, we turn to the teacher model for more information for in-domain data synthesis. In KD, the teacher model is trained on the original training data $X$, whose output is a conditional probability $T(x; \theta_t) = p(y|x, \theta_t)$, which corresponds to the confidence that $x$ belongs to $y$-th category. To align the label space, a naive method is to maximize the confidence of teacher predictions, i.e., minimizing the entropy term $H(p(y|x, \theta_t))$. However, prior works have shown that such a simple probability maximization may only lead to some ``rubbish samples''~\cite{goodfellow2014explaining} without too much useful visual information for student training. To address this problem, we propose a regularized objective to align label space , which is formalized as:
\begin{equation}
    \min_{G} \mathcal{L}_{align}(G, D, T) = \mathbb{E}_{z\sim P_z} \left[\log(1-D(C(G(z))) + H\left[p(y|G(z), \theta_t)\right] \right] \label{eqn:aligning}
\end{equation}
In Equation \ref{eqn:aligning}, the first term refers to the discrimination loss as mentioned in Equation \ref{eqn:local_gan}, which regularizes the local patterns to be authentic. The second term refers to the entropy loss for confidence maximization, which works on full images and assembles local patterns to synethsize desired categories. This objective will be simultaneously optimized with Equation (\ref{eqn:local_gan}) to keep the authenticity of local patches.

\textbf{DRO in MosaicKD.} 
As aforementioned, Equation \ref{eqn:local_gan} regularizes the local patterns to be authentic and Equation \ref{eqn:aligning} aligns the label space of synthetic data to that of training data. They collaboratively construct a new searching space for DRO framework as follows:
\begin{equation} \label{eqn:dro_hard}
    \min_{S} \max_{G} \mathcal{L}_{DRO}(G, D, S, T) = \{ \mathbb{E}_{z \sim p_z(z)}\left[ \ell_\text{KL}(T(G(z)) \| S(G(z))) \right]: \mathcal{R}(G, D, T)) \leq \epsilon \}%\lambda \mathcal{R}(G, D, T)) 
\end{equation}
where $\mathcal{R}(G, D, T)$ is a regularization term derived from Equation \ref{eqn:local_gan} and \ref{eqn:aligning}. Note that optimizing a generative adversarial networks as Equation \ref{eqn:local_gan} is equivalent to minimizing the Jensen–Shannon divergence of two patch distribution, i.e., $\ell_\text{JSD}(P^{patch}_{X^{\prime}}, P^{{patch}}_{G})$, the above regularization can be written as:
\begin{equation}
    \mathcal{R}(G, D, T) = \ell_\text{JSD}(P^{patch}_{X^{\prime}}, P^{patch}_{G}) + \mathbb{E}_{z\sim P_z}\left[H(p(y|G(z), \theta_t))\right]
\end{equation}
Regularization $\mathcal{R}(G, D, T)$ force the generator to leverage local patterns of OOD data for data synthesis, which leads to a special searching space defined on all possible schemes of patch assembling. Different to the conventional DRO, MosaicKD uses a small radius for robust optimization, where the target domain can be covered by the searching space. We relax the regularization of Equation \ref{eqn:dro_hard} to obtain an optimizable DRO objective for training, formalized as follows:
\begin{equation}\label{eqn:dro_final}
    \min_{S} \max_{G} \mathcal{L}_{DRO}(G, D, S, T) = \mathbb{E}_{x \sim P_z}\left[ \ell_\text{KL}(T(G(z)) \| S(G(z))) \right] - \lambda \mathcal{R}(G, D, T))%\lambda \mathcal{R}(G, D, T)) 
\end{equation} 

\algdef{SE}[SUBALG]{Indent}{EndIndent}{}{\algorithmicend\ }%
\algtext*{Indent}
\algtext*{EndIndent}
\definecolor{myblue}{rgb}{0.7,0.7,0.7}
\algnewcommand{\LineComment}[1]{\State \textcolor{gray}{\(\triangleright\) #1}}

\begin{algorithm}[t]
	\caption{MosaicKD for out-of-domain knowledge distillation}
	\label{alg:alg}
  \begin{flushleft}
    \textbf{Input:} Pretrained teacher $T(x; \theta_t)$, student $S(x; \theta_s)$ and out-of-domain data $X^{\prime}$. \\
    \textbf{Output:} An optimized student $S(x; \theta_s)$ 
  \end{flushleft}
	\begin{algorithmic}[1]
		%\noindent
    \State Initialize a generator $G(z;\theta_g)$ and a discriminator $D(z;\theta_d)$
    
	\Repeat
	    \LineComment{Patch Discrimination}
        \State Sample a mini-batch of OOD data $x^{\prime}$ from $X^{\prime}$ and synthetic data $x$ from $G(z)$;
        \State update discriminator to distinguish fake patches from real ones using $\mathcal{L}_{local}$ from Eqn. \ref{eqn:local_gan};

        \LineComment{Generation}
        \State Sample a mini-batch of generated data $x$ from $G(z)$;
        \State Update generator $G$ to:
        \Indent
                \State (a) fool the discriminator $D$ using $\mathcal{L}_{local}$ from Eqn. \ref{eqn:local_gan};
                \State (b) align label space with teacher $T$ using $\mathcal{L}_{align}$ from Eqn. \ref{eqn:aligning};
                \State (c) fool the student $S$ using $\mathcal{L}_{DRO}$ from Eqn. \ref{eqn:dro_final};
        \EndIndent
        
        \LineComment{Knowledge Distillation}
        \For{$j$ steps}:
        \State Sample generated samples from $G(z)$;
    	\State Update student through knowledge distillation using $\mathcal{L}_{DRO}$ from Eqn. \ref{eqn:dro_final}
    	\EndFor
	\Until converge
	\end{algorithmic}
\end{algorithm}

\subsection{Optimization}

\textbf{Patch Discriminator.} For training efficiency, the discriminator in equation \ref{eqn:local_gan} can be implemented as a Patch GAN~\cite{isola2017image} with carefully designed receptive fields and patch overlap. Specifically, we stack several convolutional layers to build a fully convolutional network, whose output is a score map instead of a single true-or-fake scalar. Each score unit accepts a $L\times L$ local patches for discrimination. We apply an additional stride downsampling with step size $s$ on the score map to control the overlap between patches. A large step size $s$ will lead to more independent patches, which effectively reduce the structure restrictions in OOD images. 

\textbf{Full Algorithm.} The full algorithm of MosaicKD is summarized in Alg. \ref{alg:alg}, where a generator $G(z; \theta_g)$, a discriminator $D(x; \theta_d)$, a fixed teacher model $T(x; \theta_t)$ and a student $S(x; \theta_s)$ are collectively optimized under the guidance of $\mathcal{L}_{local}$, $\mathcal{L}_{align}$ and $\mathcal{L}_{DRO}$.

\section{Experiments}

\subsection{Settings}

\textbf{Datasets.} The proposed method is evaluated on two mainstream vision tasks, \textit{i.e.}, image classification and semantic segmentation. Four datasets are considered in our experiments as in-domain training set, including CIFAR-100~\cite{krizhevsky2009learning}, CUB-200~\cite{WelinderEtal2010}, Stanford Dogs~\cite{KhoslaYaoJayadevaprakashFeiFei_FGVC2011} and NYUv2~\cite{Silberman:ECCV12}. For OOD-KD, we substitute original training data with OOD data, including CIFAR-10~\cite{krizhevsky2009learning}, Places365~\cite{zhou2017places}, ImageNet~\cite{deng2009imagenet} and SVHN~\cite{Netzer2011ReadingDI}. 

\textbf{Evaluation metrics.} For image classification, accuracy and Frechet Inception Distance (FID) are used to evaluate different methods. FID indicates the divergence of two datasets, which was originally used to assess the synthesis quality of GANs~\cite{heusel2017gans}:
\begin{equation}
  FID = \| \mu_1 - \mu_2 \|_2^2 + tr \left(\Sigma_1+\Sigma_2-2\left(\Sigma_1\Sigma_2\right)^{1/2} \right)
\end{equation}
where $(\mu_1, \Sigma_1)$ and $(\mu_2, \Sigma_2)$ are the mean and covariance statistics of generated and original samples. For semantic segmentation, we use mean Intersection of Unions (mIoU) as the performance metric. More details about datasets, training protocol, and metrics can be found in supplementary materials.

\subsection{Results of Knowledge Distillation}
\newcommand{\std}[1]{\scriptsize $\pm${#1}}

\def\scoreup#1{$(\color{mygreen} \uparrow #1)$}
\def\scoredown#1{$(\color{red} \downarrow #1$)}

\begin{table*}[t]
  \centering
  \resizebox{\textwidth}{!}{
  \begin{tabular}{l c c c c c c c}
      \toprule
      \bf \multirow{2}{*}{Method} & \bf \multirow{2}{*}{Data} & \bf resnet-34  & \bf vgg-11 & \bf wrn40-2  & \bf wrn40-2 & \bf wrn40-2 & \bf \multirow{2}{*}{Average} \\
                                  &                           & \bf resnet-18  & \bf resnet-18 & \bf wrn16-1  & \bf wrn40-1 & \bf wrn16-2 &  \\                  
    \hline  
      Teacher & \multirow{3}{*}{\shortstack{CIFAR-100\\(Original Data)}} & 78.05 & 71.32  & 75.83 & 75.83   & 75.83 & 75.37 \\ % lr 0.1
      Student &                       & 77.10 & 77.10      & 65.31 & 72.19   & 73.56  & 73.05 \\
      KD~\cite{hinton2015distilling}   &  & 77.87 & 75.07 & 64.06 & 68.58 & 70.79 & 72.27\\
      \hline
      DAFL~\cite{chen2019data} & \multirow{4}{*}{Data-Free} & 74.47   & 54.16 & 20.88  & 42.83 & 43.70 & 47.20  \\
      ZSKT~\cite{micaelli2019zero} & & 67.74   & 54.31  & 36.66 & 53.60  & 54.59 & 53.38  \\
      DeepInv.~\cite{yin2019dreaming} & & 61.32   & 54.13  & 53.77  & 61.33  & 61.34 & 58.38 \\
      DFQ~\cite{choi2020data} & & 77.01   & 66.21  & 51.27 & 54.43  & 64.79 & 62.74 \\
      %GDFD~\cite{luo2020largescale} & & 77.02   & - & - & - & - & - \\
      %CMI~\cite{fang2021contrastive} & & \bf 77.04   & 70.56  & 57.91  & 68.75  & 68.88 & 68.62  \\
      \hline

      KD~\cite{hinton2015distilling} & \multirow{7}{*}{\shortstack{CIFAR-10\\(OOD Data)}} & 73.55  & 68.04 & 47.47 & 61.17 & 63.48 & 62.74\\
      Balanced~\cite{nayak2021effectiveness} & & 68.54 & 64.14 & 50.50  & 56.50  & 57.33 & 59.40  \\
      FitNet~\cite{romero2014fitnets} & & 70.14  & 67.52  & 50.31  & 60.17  & 60.60 & 63.15 \\
      RKD~\cite{park2019relational} & & 67.45  & 63.06  & 45.37  & 53.29  & 57.10 & 57.25 \\
      CRD~\cite{tian2019contrastive} & & 71.23 & 66.48 & 47.00 & 59.59 & 61.37 & 61.13 \\
      SSKD~\cite{xu2020knowledge} & & 73.81 & 68.72 & 49.57 & 60.71 & 64.61 & 63.48 \\
      Ours & & \bf 77.01  & \bf 71.56  & \bf 61.01  & \bf 69.14  & \bf 69.41 & \bf 69.55  \\
      \arrayrulecolor{gray!50}\hline
      
      KD~\cite{hinton2015distilling} & \multirow{5}{*}{\shortstack{ImageNet$^\dagger$\\(OOD Subset)}}& 50.89  & 50.52  & 36.54   & 36.87  & 41.69 & 43.30  \\
      Balanced~\cite{nayak2021effectiveness} & & 41.74   & 47.04  & 31.61   & 29.57  & 35.65 & 37.12  \\
      FitNet~\cite{romero2014fitnets} & & 60.15  & 58.23  & 42.63  & 44.21  & 48.53 & 50.75  \\
      RKD~\cite{park2019relational} & &  40.26   & 35.80  & 31.15   & 24.95   & 34.48 & 33.32   \\
      Ours & & \bf 75.81  & \bf 68.94  & \bf 59.32   &  \bf 66.61   &  \bf 67.36 & \bf 67.60  \\
      \arrayrulecolor{gray!50}\hline
      
      KD~\cite{hinton2015distilling} & \multirow{5}{*}{\shortstack{Places365$^\dagger$\\(OOD Subset)}} & 43.49  & 46.24  & 33.28  & 31.39   & 36.37 & 38.15 \\
      Balanced~\cite{nayak2021effectiveness} & & 28.16  & 38.85   & 23.22  & 21.54  & 28.62 & 28.08 \\
      FitNet~\cite{romero2014fitnets} & & 54.08   & 54.15   & 36.33  & 44.21  & 38.74 & 45.50   \\
      RKD~\cite{tian2019contrastive} & & 30.25  & 33.06  & 28.07  & 21.12  & 21.12 & 26.72  \\
      Ours & & \bf 74.70  & \bf 68.55  & \bf 56.70  & \bf 65.34   & \bf 65.89 & \bf 66.23  \\
      \arrayrulecolor{gray!50}\hline
      
      KD~\cite{hinton2015distilling} & \multirow{5}{*}{\shortstack{SVHN\\(OOD Data)}} & 31.55  & 34.00  & 19.77  & 23.07  & 24.75 & 26.63  \\
      Balanced~\cite{nayak2021effectiveness} & & 26.93 & 29.34  & 16.18  & 18.96  & 21.50 & 22.58   \\
      FitNet~\cite{romero2014fitnets} & & 33.69  & 36.22  & 20.02  & 23.72  & 25.41 & 27.81   \\
      RKD~\cite{park2019relational} & & 26.83  & 27.31  & 18.09  & 22.55  & 24.29 & 23.81  \\
      Ours & & \bf 47.18  & \bf 37.63  & \bf 31.87  & \bf 45.84  & \bf 44.40 & \bf 41.38   \\
      \arrayrulecolor{black}\hline
  \end{tabular} 
  }
  \vspace{-2mm}
  \caption{Test accuracy (\%) of student networks trained with the following settings: conventional KD with original training data, data-free KD with synthetic data, and OOD-KD with OOD data. $\dagger$: As Places365 and ImageNet contain some in-domain samples, we craft OOD subsets with low teacher confidence (high entropy) from the original dataset, so as to match our OOD setting.  }
  \vspace{-2mm}
  \label{tbl:benchmark_classification}
\end{table*} 

\begin{table}[t]
\begin{tabularx}{\textwidth}{*{2}{>{\centering\arraybackslash}X}}
  \begin{tabular}{l c c c}
      \toprule
      \bf Method & \bf Data & \bf FLOPs & \bf mIoU \\
      \hline  
      Teacher & \multirow{2}{*}{NYUv2}  & 41G & 0.519 \\
      Student &  & 5.54G & 0.375 \\
      \hline
      ZSKT    & \multirow{2}{*}{Data-Free} & 5.54G & 0.364 \\
      DAFL    &   & 5.54G & 0.105 \\
      %DeepInversion &  & 5.54G  &  \\
      \hline
      KD & \multirow{2}{*}{ImageNet} & 5.54G & 0.406 \\
      %DFND & & & 0.378 \\
      Ours & & 5.54G & \bf 0.454 \\
      \hline
    \end{tabular} 
    \vspace{2mm}\label{tbl:newv2_seg}
\captionof{table}{Mean Intersection over Union (mIoU) of student models on NYUv2 data set.} %The teacher is a deeplabv3-resnet50 model, while the student is lightweight a deeplabv3-mobilenet model.} \label{tbl:newv2_seg}
&     
    \iffalse
    \begin{tabular}{l c c c}
     \toprule
     \bf \multirow{2}{*}{Method} & \bf Res50 & \bf Res50 & \bf Res50 \\
     & \bf Res18 & \bf Res50 & \bf Mobv2 \\
     \hline  
     Scratch & 75.45 & 68.45 & 70.01 \\
     DeepInversion & 68.00 & -- & -- \\
     LS-GDFD & 69.75 & 54.66  & 43.15 \\
     Places365+KD & 55.74 & 45.30 & 39.89 \\
     Places365+Ours &  &  \\
     \hline
  \end{tabular} 
  \fi
  
  \begin{tabular}{l c c}
     \toprule
     \bf Method & \bf CUB-200 & \bf Stanford Dogs \\
     \hline  
     Teacher & 49.41 & 56.65 \\
     Student & 41.44 & 48.61 \\
     \hline
     KD          & 11.07  &   10.24   \\
     Balanced    & 4.56   &   6.42    \\
     FitNet      & 18.12  &   19.13    \\
     Ours        & \bf 26.11 &  \bf 28.02        \\
    \hline
  \end{tabular} 
\vspace{2mm}
\captionof{table}{Test accuracy of student networks on fine-grained datasets.}\label{tbl:fine-grained}
\end{tabularx}
\vspace{-8mm}
\end{table}

\textbf{CIFAR-100.} Table \ref{tbl:benchmark_classification} reports the results of knowledge distillation on CIFAR-100 dataset. Here we use CIFAR-10, ImageNet, Places365 and SVHN as OOD data to evaluate MosaicKD for OOD settings. We compare the proposed MosaicKD to various baselines, including data-free KD methods~(DAFL~\cite{chen2019data}, ZSKT~\cite{micaelli2019zero}, DeepInv.~\cite{yin2019dreaming}, DFQ~\cite{choi2020data}) and OOD-KD methods naively adapted from state-of-the-art KD approaches~(BKD~\cite{hinton2015distilling}, Balanced~\cite{nayak2021effectiveness}, FitNet~\cite{romero2014fitnets}, RKD~\cite{park2019relational}, CRD~\cite{tian2019contrastive} and SSKD~\cite{xu2020knowledge}).

As shonw in Table \ref{tbl:benchmark_classification}, despite the mismatched distributions, conventional KD approaches still learn some useful but incomplete knowledge from OOD data~(\textit{i.e.}, yielding significantly superior performance to random guessing), which indicates the existence of shared patterns between OOD data and training data. Further, some exploration was taken to evaluate importance of category balance and representation transfer for OOD-KD. First, we balance the OOD data by re-sampling the scarce categories according to teacher's predictions. However, results show that balancing the OOD data may not help students learn correct class information, because most samples in OOD data are just misclassified outliers. In the context of OOD settings, the balance operation may lead to over-fitting on outliers, which may further degrade the student performance on the test set. As mentioned before, OOD data and original data may share some local patterns, which can be extracted by shallow layers of networks. We apply four representation transfer approaches, i.e., FitNet, RKD, CRD and SSKD to study the their role in OOD-KD. Compared with RKD that focus instance relation, we found that response-based methods like fitnet can transfer more helpful information in OOD settings, where the student directly imitates the teacher's intermediate outputs of teachers. In general, transferring low-level features sometimes can be helpful for OOD-KD. However, note that CRD works on the high-level representation extracted from the penultimate layer, transferring these knowledge may be inappropriate for OOD-KD because high-level features may be unrelated to target tasks. 

In this work, we handles the OOD-KD problem as a generative problem, instead of directly using OOD data for training. The proposed method leverage local patterns of OOD data for data synthesis, where some task-related patterns will be ``assembled'' from shared local patches. Results show that these re-assembled data can effectively transfer knowledge from teachers to students. In Table \ref{tbl:benchmark_classification}, we also extend our method to different types of OOD sets. We found that the performance of MosaicKD is related to the degree of domain divergence between OOD data and original data. For example, ImageNet is an object recognition dataset while Places365 is a scene classification dataset. Results show that, for the target data CIFAR-100, MosaicKD can achieve better performance on ImageNet compared to Places365. 

\textbf{Fine-grained Classification.} To further study the effectiveness of our approach, we conduct knowledge distillation on fine-grained datasets as shown in Table \ref{tbl:fine-grained}, using Places365 as OOD data. OOD-KD on fine-grained data is a challenging problem, because different categories is visually similar. Results show that our method achieves superior performance compared to baselines methods.

\textbf{Semantic Segmentation.}
Semantic segmentation can also be viewed as a classification task, where the network is trained to predict the category of each pixel. We apply our method to the NYUv2 dataset, following the protocol in \cite{fang2019data}. The teacher network is a deeplab v3 network with resnet-50 backbone. The student is a freshly initialized deeplabv3-MobileNetv2 model. Our method can effectively improve the knowledge transfer on OOD data and achieve competitive results even compared to vanilla KD settings.

\begin{figure}[t]
  \begin{center}
     \includegraphics[width=\linewidth]{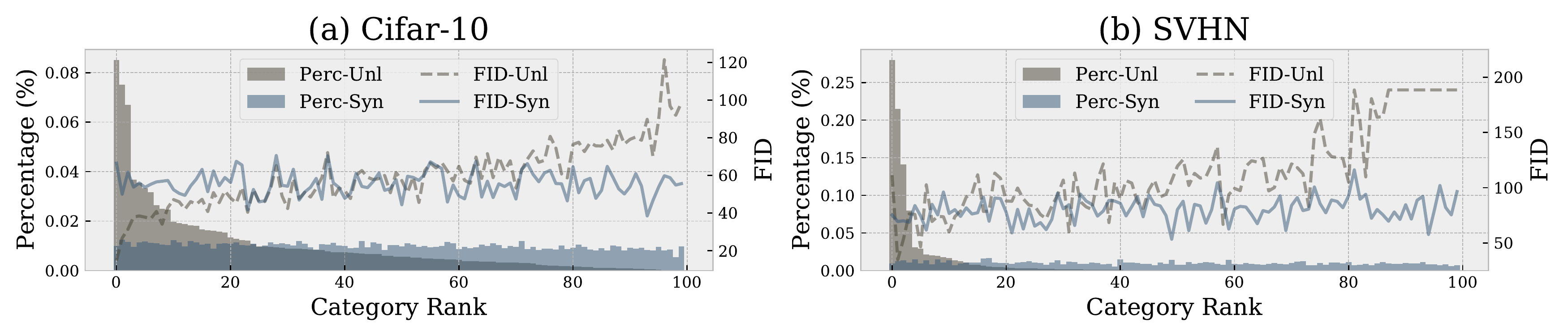}
     
  \end{center}\vspace{-3mm}
     \caption{Statistical information of OOD data and out generated data. Category percentage (the first y-axis) and FID score (the second y-axis) to original data (CIFAR-100) is reported.} \label{fig:cls_balance}
\end{figure}

\subsection{Quantitative Analysis}

\textbf{Data balance and FID.} Figure \ref{fig:cls_balance} provides some statistical information of OOD data and generated samples, including the category balance predicted by teachers and the per-class FID scores. The category is ranked according to their percentages. Note that the original CIFAR-10 dataset only contains 10 categories, which is very limited compared with the 100 categories of CIFAR-100. As illustrated in Figure \ref{fig:cls_balance} (a), some CIFAR-100 categories are missing in CIFAR-10. Besides, the large FID between OOD data and original training data also indicates that, even though some samples are categorized to some classes by the teacher, they may still belong to outliers. by contrast, our method successfully balances different CIFAR-100 categories and alleviates the domain gap (lower class FID), especially for unbalanced categories.

\begin{table*}[t]
  \centering
  \small
  \begin{tabular}{l c c c c c c c c}
     \toprule
     \multicolumn{2}{c}{\bf Patch size} &  \bf 1 & \bf 2 & \bf 4 & \bf 8 & \bf 18 & \bf 22 & \bf 32 \\
     \hline  
     \multirow{2}{*}{CIFAR-10} & Acc.  & 52.22 & 55.58 & 58.94  & 61.01 & \bf 61.03 & 58.94 & 51.34 \\
    & FID & 2.43 & 4.47 & 8.78 & 16.82 & 22.81 & 26.07 &  28.30 \\
    \hline
    \multirow{2}{*}{Places365} & Acc.  & 46.44 & 45.90 & 50.67 & \bf 56.70 & 53.99 & 53.44 & 40.29  \\
    & FID &  12.32 &  14.77 &  21.32 &  30.09 &  35.54 & 38.64 & 41.41 \\
     \hline
    \multirow{2}{*}{SVHN} & Acc.  & 19.83 & 21.86 & \bf 32.09 &  31.87 & 21.05  & 22.08 & 20.54 \\
    & FID & 93.39 & 146.76  & 145.64  & 143.72 & 148.33 & 147.25 & 148.94  \\
    \hline
  \end{tabular} 
  \caption{Test accuracy (\%) of students obtained with different patch sizes. The Patch FID score between OOD data and original data is also reported. Results show that our approach requires smaller patch sizes to handle severe domain discrepancies.} \label{tbl:patch} \vspace{-3mm}
\end{table*}

\begin{figure*}[t]
  \begin{center}
     \includegraphics[width=\linewidth]{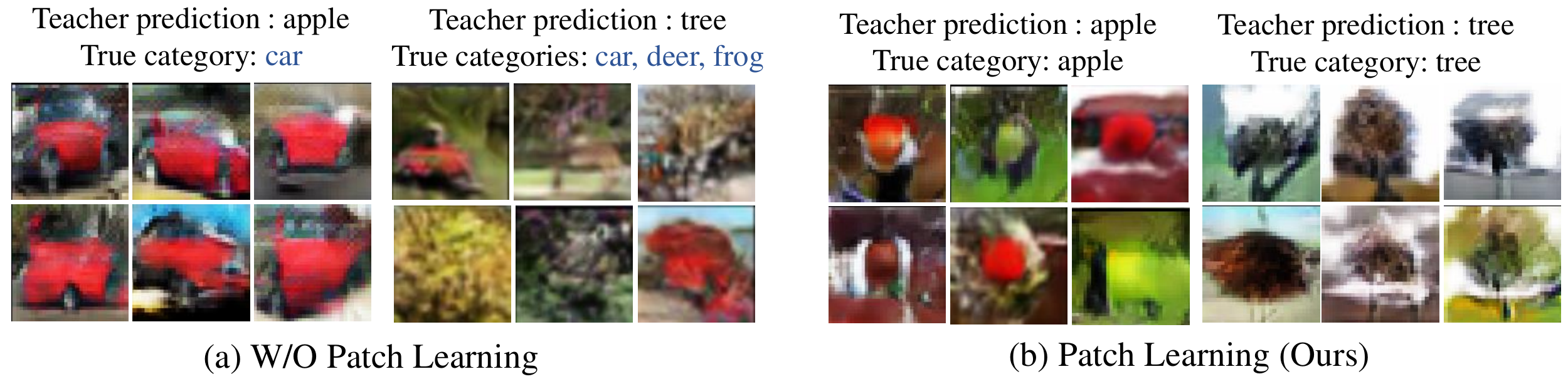}
  \end{center}\vspace{-3mm}
     \caption{Visualization of synthetic data with and without patch learning. GANs without patch learning will be trapped by OOD data and fails to present correct semantic for different categories (highlighted in blue). In our method, the semantic can be correctly aligned with target domain.} \label{fig:vis_patch_learning}
\end{figure*}

\begin{figure}[t]
\begin{tabularx}{\textwidth}{*{2}{>{\centering\arraybackslash}X}}
  \small
  \begin{tabular}{l c c c}
     \toprule
     \bf \multirow{2}{*}{Method} & \bf wrn40-2 & \bf wrn40-2 & \bf wrn40-2 \\
     & \bf wrn16-1 & \bf wrn40-1 & \bf wrn16-2 \\
     \hline  
     ours  & \bf 61.01 & \bf  69.14 & \bf 69.41 \\
     w/o patch & 51.34 & 56.23 & 57.23 \\
     w/o disc. & 44.05 & 58.11 & 59.73 \\
     w/o adv.  & 55.11 & 67.57 & 68.25\\
     \hline
  \end{tabular} 
  %\vspace{-3mm}
\captionof{table}{Test accuracy (\%) of students in different ablation settings.} \label{tbl:ablation}
&     
\small
\includegraphics[width=\linewidth,valign=m]{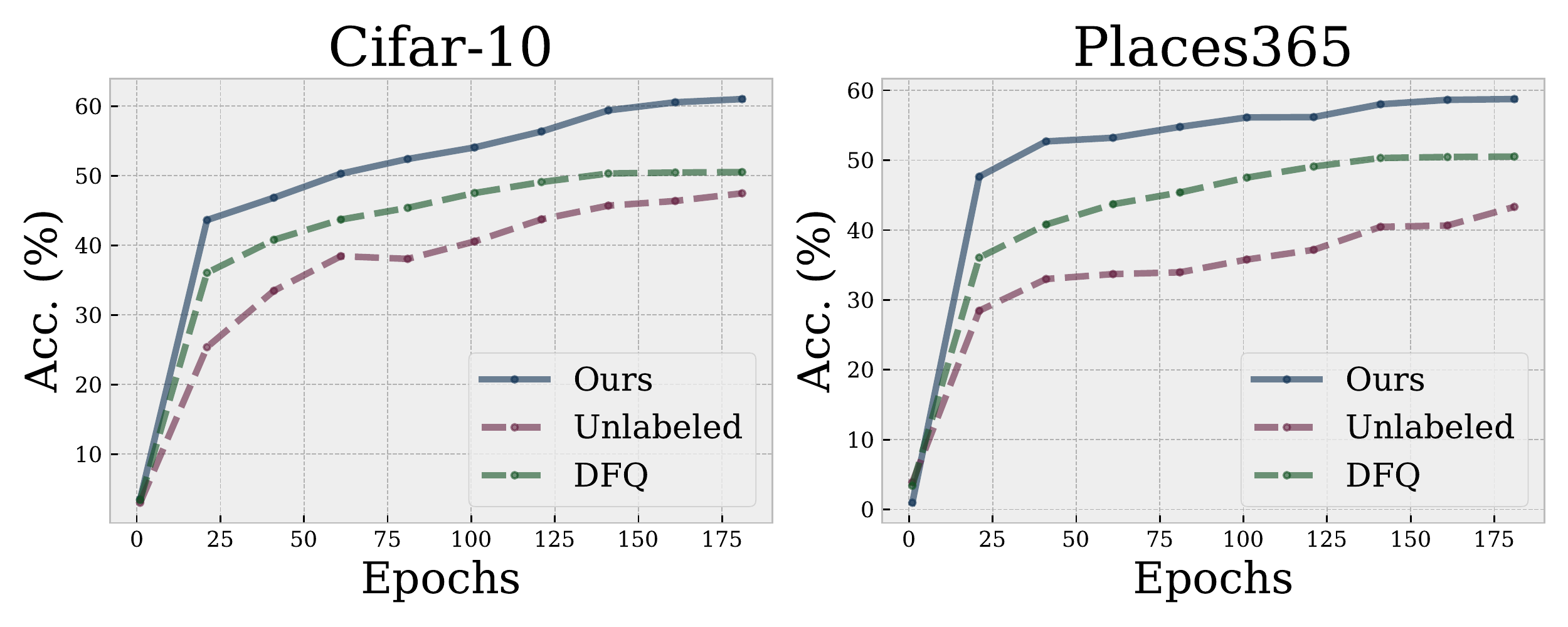} 
\vspace{-2mm}
\captionof{figure}{Training efficiency of our method compared with DFQ and vanilla KD.}\label{fig:loss_curve}
\end{tabularx}
\end{figure}

\textbf{The influence of patch size.} Patch size plays an essential role in our method, which determines the flexibility of data synthesis. The optimal patch size actually depends on the divergence between OOD data and in-domain data. For OOD data with large domain discrepancy, small patches are usually required due to the limited local similarity. As shown in Table \ref{tbl:patch}, we evaluate our method with different patch sizes and report the test accuracy of student models as well as the patch FID. According to the Table \ref{tbl:patch}, we find that for OOD data like CIFAR-10 and Places365, a large patch size (e.g., 18) can be used for student learning. However, for SVHN dataset, a smaller patch size would be more appropriate, as SVHN severely diverges from CIFAR-100.

\textbf{Ablation study.} In this section, we conduct ablation to understand the role of patch learning further. We consider the following settings: (a) MosaicKD (b) mosaicKD without patch learning (c) mosaicKD without discriminator (d) mosaicKD wihout adversarial training. As shown in Table \ref{tbl:ablation}, we find that full image discrimination sometimes even lead to worse results than MosaicKD without discrimination. Figure \ref{fig:vis_patch_learning} visualizes the synthetic data with or without patch learning. Results show that the generator without patch learning is trapped by the label space of OOD data, failing to synthesize in-domain categories like trees and apples. %by contrast, our method can produce desired objects for knowledge distillation. 
\section{Discussion}
\textbf{Relation to unlabeled knowledge distillation.} In the literature of knowledge distillation, a slice of works also use unlabeled data for student learning. However, they either make an i.i.d. assumption about unlabeled data and original training~\cite{ye2019student}, or assume that there are sufficient in-domain samples inside unlabeled set~\cite{xu2019positive}. In this work, we allow the unlabeled set to be fully OOD, which is more challenging than existing unlabeled settings.

\textbf{Relation to data-free knowledge distillation.} The commonality between the data-free algorithm and MosaicKD lies in that they both solve the KD problem through data synthesis. However, data-free KD leverages some simple priors such as category confidence~\cite{chen2019data} and gaussian assumptions~\cite{yin2019dreaming}, which ignores the structural details in natural images. By contrast, MosaicKD achieves data synthesis in an  assembling-by-dismantling manner, where natural patterns can be utilized to improve synthesis quality. The training curves of different methods can be found in Figure \ref{fig:loss_curve}.

\section{Conclusion}
In this work, we propose a novel approach, termed
as MosaicKD, that enables us to take advantage
of only the OOD data for KD. MosaicKD follows 
a assembling-by-dismantling scheme,
where a synthetic sample is generated 
by locally resembling the real-world OOD data
while globally fooling the pre-trained teacher. 
This is technically achieved 
through a handy yet effective four-player
min-max game, in which 
a generator, a discriminator,
and a student network
are learned
in the presence of 
a pre-trained teacher.
We validate MosaicKD over classification 
and semantic segmentation tasks across various benchmarks,
and showcase that it yields results
significantly superior
to the state-of-the-art techniques 
on OOD data.

\clearpage
\section{Acknowledgements and Disclosure of Funding}
This work is supported by National Natural Science Foundation of China (U20B2066, 61976186), Key Research and Development Program of Zhejiang Province (2020C01023), the Major Scientific Research Project of Zhejiang Lab (No. 2019KD0AC01), the Fundamental Research Funds for the Central Universities, Alibaba-Zhejiang University Joint Research Institute of Frontier Technologies, 
Start-Up Grant from National University of Singapore (R-263-000-E95-133),
and MOE AcRF TIER~1 FRC Research Grant (R-263-000-F14-114).

{
  \small
  \bibliographystyle{plain}
  \bibliography{mosaic_kd}
}

\clearpage

\section{Appendix}

In this document, we provide details
and supplementary materials that 
cannot fit into the main manuscript due to 
the page limit. Specifically,
we provide optimization details of MosaicKD in Sec.~\ref{sec:optimization}, experimental settings in Sec. \ref{sec:exp_settings}, and more experimental results in Sec. \ref{sec:exp_results}.

\subsection{Optimization Details}\label{sec:optimization}

\subsubsection{Alleviating Mode Collapse.} 

In this work, we deploy a generator to
synthesize the transfer set for knowledge distillation. 
Nevertheless,  GANs are known to
suffer from mode collapse and 
fail to produce diverse patterns. 
To this end, we leverage both OOD data and synthetic ones to train our student models,
so that the generator does not need to synthesize all samples for KD. Besides, an additional balance loss is deployed to alleviate mode collapse during training, defined as:
\begin{equation}
    L_{balance} = -H(\mathbb{E}_{x\sim P_G}(p(y|x, \theta_t))) \label{eqn:balance}
\end{equation}
where $p(y|x, \theta_t)$ is the probability prediction after softmax, and $P_G$ denotes the distribution of generated samples. Minimizing Eq.~(\ref{eqn:balance}) will enforce the class to be balanced during the synthesizing process. 

\subsubsection{Objectives of MosaicKD.} As shown in the main manuscript, MosaicKD aims to solve a distributionally robust optimization (DRO) problem as follows:
\begin{equation}
    \min_{S} \max_{G} \{ \mathbb{E}_{x \sim P_{G}}\left[ \ell_\text{KL}(T(x; \theta_t) \| S(x; \theta_s)) \right]: \mathcal{R}(G, D, T)) \leq \epsilon \}%\lambda \mathcal{R}(G, D, T)) 
\end{equation}
\noindent where $\mathcal{R}(G, D, T)) \leq \epsilon$ defines the search space, i.e., a ball space with radius $\epsilon$ centered at an distribution satisfying  $\mathcal{R}(G, D, T))=0$. The specific form of center distribution is unknown, but we can still train a generator $G$ to approximate it. Note that Eq.~(\ref{eqn:dro_hard}) is intractable due to the non-differentiable condition on the search space. With the help of lagrange duality, we can re-express the inner part of 
Eq.~(\ref{eqn:dro_hard}) as  follows:
\begin{equation}\label{eqn:dro_derivation}
\begin{split}
& \max_{G} \{ \mathbb{E}_{x \sim P_{G}}\left[ \ell_\text{KL}(T(x; \theta_t) \| S(x; \theta_s)) \right]: \mathcal{R}(G, D, T)) \leq \epsilon \} \\
& = \max_G \min_{\lambda\ge0} \{ \mathbb{E}_{x \sim P_{G}}\left[ \ell_\text{KL}(T(x; \theta_t) \| S(x; \theta_s)) \right] + \lambda \cdot (\epsilon-\mathcal{R}(G, D, T)))\}\\
& \leq \min_{\lambda\ge 0} \max_G \{\lambda\epsilon + \mathbb{E}_{x \sim P_{G}}\left[ \ell_\text{KL}(T(x; \theta_t)\| S(x; \theta_s)) \right] - \lambda\cdot\mathcal{R}(G, D, T)) \} \\
& = \min_{\lambda\ge 0} \{ \lambda\epsilon + \max_G \{\mathbb{E}_{x \sim P_{G}} \left[ \ell_\text{KL}(T(x; \theta_t) \| S(x; \theta_s)) \right] - \lambda\cdot\mathcal{R}(G, D, T)) \} \}
\end{split}
\end{equation}

\noindent where $\lambda$ is Lagrangian multiplier and $\lambda\epsilon$ is a constant term. If $\mathcal{R}(G, D, T))\leq\epsilon$, we choose $\lambda=0$, i.e., no restriction on $\mathcal{R}(G, D, T))$, to obtain the minimal cost. If $\mathcal{R}(G, D, T))>\epsilon$, then a large $\lambda$ should be applied as a penalization. According to the derivation of Eq.~(\ref{eqn:dro_derivation}), we obtain a relaxed version of the intractable Eq.~(\ref{eqn:dro_hard}), expressed as follows:
\begin{equation}
    \min_{S} \max_{G} \mathcal{L}_{DRO}(G, D, S, T) = \mathbb{E}_{x \sim P_{G}}\left[ \ell_\text{KL}(T(x; \theta_t), S(x; \theta_s)) \right] - \lambda \mathcal{R}(G, D, T))%\lambda \mathcal{R}(G, D, T)) 
\end{equation} 

\subsubsection{GAN Training and JS Divergence.}

Following the conventions of prior works,
we write the GAN training objective as follows, 
\begin{equation}\label{eqn:gan}
    \min_G \max_D V(D, G) = \mathbb{E}_{x\sim P_{data}}\left[\log D(x)\right] + \mathbb{E}_{z\sim P_z}\left[log(1-D(G(z)))\right].
\end{equation}

As proposed in \cite{goodfellow2014generative}, for a fixed generated $G$ 
and a given data distribution $P_{data}$, the optimal discriminator $D$ is achieved when
\begin{equation}
    D^{*}(x) = \frac{P_{data}(x)}{P_{data}(x) + P_G(x)}
\end{equation}
We then replace the discriminator in Eq.~(\ref{eqn:gan}) with the optimal one $D^{*}$, which leads to the following optimization for generator $G$:

\begin{equation}
  \begin{split}
    \min_G V(G, D^{*}) & = \mathbb{E}_{x\sim P_{data}}\left[\log D^{*}(x)\right] + \mathbb{E}_{z\sim P_z}\left[log(1-D^{*}(G(z)))\right] \\
    & = \mathbb{E}_{x\sim P_{data}}\left[\log D^{*}(x)\right] + \mathbb{E}_{x\sim P_G}\left[log(1-D^{*}(x))\right] \\
    & = \mathbb{E}_{x\sim P_{data}}\left[\log \frac{P_{data}(x)}{P_{data}(x) + P_G(x)} \right] + \mathbb{E}_{x\sim P_G}\left[log(\frac{P_{G}(x)}{P_{data}(x) + P_G(x)})\right] \\
    & = -log(4) + \ell_{\text{KL}}(P_{data}\|\frac{P_{data}+P_G}{2}) + \ell_{\text{KL}}(P_{G}\|\frac{P_{data}+P_G}{2}) \\
    & = -log(4) + 2\cdot \ell_{\text{JSD}}(P_{data}\|P_G)
  \end{split}
\end{equation}

Therefore, as mentioned in the manuscript, we optimize generative adversarial networks to minimize the regularization term $R(G,D,T)$, which is equivalent to optimizing the JS divergence between patch distributions.

\subsection{Experimental Settings}\label{sec:exp_settings}

\begin{wrapfigure}{R}{0.5\textwidth} \label{alg:ood_subset}
\vspace{-6mm}
    \begin{minipage}{0.5\textwidth}
      \begin{algorithm}[H]
      \begin{flushleft}
        \textbf{Input:} dataset $D$, Pretrained teacher $T(x; \theta_t)$, \\
        \textbf{Output:} OOD subset $D^{\prime}$
      \end{flushleft}
        \caption{OOD subset selection}
        \begin{algorithmic}[1]
          \State $H \leftarrow []$
          \For{$x_i$ in D}:
             \State obtain prediction $p(y|x_i)=T(x)$
             \State calculate the entropy $h_i = H(p(y|x_i))$
             \State $H$.append($h_i$)
          \EndFor
          \State index $\leftarrow$ topk-index($H$);
          \State $D^{\prime} \leftarrow D[\text{index}]$;
          \State return $D^{\prime}$
        \end{algorithmic}
      \end{algorithm}
    \end{minipage}
  \end{wrapfigure}
  
\paragraph{Datasets.} The proposed method is evaluated on two mainstream vision tasks, \textit{i.e.}, 
image classification and semantic segmentation, over four labeled datasets for teacher training and four OOD data for student learning, as summarized in Table \ref{tbl:datasets}. Note that 
CIFAR-100, ImageNet, and Places365 may contain in-domain categories. 
We craft OOD subset from the full ImageNet and Places365 datasets by selecting samples with low prediction confidence, as described in Algorithm \ref{alg:ood_subset}. These OOD subsets can be viewed as out-of-domain data for CIFAR-100. Besides, we resize the OOD data to the same resolution as in-domain data, e.g., $32\times 32$ for CIFAR-100, $64\times64$ for fine-grained datasets, and $128 \times 128$ for NYUv2.

\paragraph{Network Training.} In this work, all teacher models are trained using the in-domain datasets listed in Table \ref{tbl:datasets} with cross entropy loss. We use SGD optimizer with $\{lr=0.1, weight\_decay=1e-4, momentum=0.9\}$ and train each model for 200 epochs, with cosine annealing scheduler. In knowledge distillation, student models are crafted using unlabeled datasets, where only the soft targets from teachers are utilized. We use the same training protocols as the teacher training and report the best student accuracy on test sets. We use Adam for optimization, with hyper-parameters $\{lr=1e-3, \beta_1=0.5, \beta_2=0.999\}$ for the
generator and discriminator.

\begin{table}[h]
\centering
    \begin{tabular}{l c c c c} 
     \toprule
     \bf In-Domain Data & \bf Training & \bf Testing & \bf Num. Classes \\
     \hline  
     CIFAR-100 & 50,000 & 10,000 & 100   \\
     CUB200 & 5,994 & 5,794 & 200 \\
     Stanford Dogs & 12,000 & 8,580 & 120 \\
     NYUv2 & 795 & 654 & 13 \\
     \toprule
     \bf OOD Data & \bf Training & \bf Testing & \bf Num. Classes \\
     \hline  
     CIFAR-10 & 50,000 & 10,000 & 100   \\
     ImageNet-OOD & 50,000 & - & -  \\
     Places365-OOD & 50,000 & - & - \\
     SVHN & 73,257 & 26,032 & 10 \\
     ImageNet & 1,281,167 & 50,000 & 1000 \\
     Places365 & 1,803,460 & 36,500 & 365 \\
     \hline
     \end{tabular}
     \vspace{2mm}
     \caption{Statistical information of in-domain and out-of-domain datasets} \label{tbl:datasets}
\end{table}

\paragraph{Generator and Discriminator.} The architecture of GAN for CIFAR-100 dataset is illustrated in Tables \ref{tbl:generator} and \ref{tbl:discriminator}. For  CUB-200 ($64\times 64$) and NYU ($128\times 128$), we add more convolutional layers and upsampling or sampling layers to generate high-resolution images. 

\begin{table}[t]
  \begin{tabularx}{\textwidth}{*{2}{>{\centering\arraybackslash}X}}
    \begin{tabular}{c}
        \toprule
        Input: $z \in \mathbb{R}^{100} \sim \mathcal{N}(0, I)$ \\
        \hline
        $\text{Linear(100)} \rightarrow 8 \times 8 \times 128$ \\
        Reshape, BN, LeakyReLU \\
        $\text{Upsample} 2\times$ \\
        $3\times 3 ~\text{Conv} 128 \rightarrow 128, ~\text{BN, LeakyReLU}$ \\
        $\text{Upsample} 2\times$ \\
        $3\times 3 ~\text{Conv} 128 \rightarrow 64, ~\text{BN, LeakyReLU}$ \\
        $3\times 3 ~\text{Conv} 64 \rightarrow 3, ~\text{Sigmoid}$ \\
       \hline 
      \end{tabular} 
      \vspace{2mm}
  \captionof{table}{Generator archicture for CIFAR-100. We add more convolutional layers and upsample layers for datasets with larger resolution.} \label{tbl:generator}
  &     
    \begin{tabular}{c}
        \toprule
        Input: $x \in \mathbb{R}^{32\times32\times3}$ \\
        \hline
        $3\times 3 ~\text{Conv} 3 \rightarrow 64, ~\text{stride}=2$ \\
        BN, LeakyReLU \\
        $3\times 3 ~\text{Conv} 64 \rightarrow 128, ~\text{stride}=2$ \\
        BN, LeakyReLU \\
        $3\times 3 ~\text{Conv} 128 \rightarrow 1, ~\text{stride}^{\dagger}=1$ \\
        Sigmoid \\
       \hline 
      \end{tabular} 
      \vspace{2mm}
  \vspace{2mm}
  \captionof{table}{Patch Discriminator archicture for CIFAR-100. $\dagger$: The final stride controls the patch overlap of MosaicKD.} \label{tbl:discriminator}
  \end{tabularx}
  \vspace{-8mm}
  \end{table}

\subsection{More Experimental Results}\label{sec:exp_results}

\begin{table}[b]
    \centering
    \begin{tabular}{l c c c}
     \toprule
     \bf \multirow{2}{*}{Stride} & \bf wrn40-2 & \bf wrn40-2  & \bf wrn40-2 \\
     & \bf wrn16-1 & \bf wrn40-1 & \bf wrn16-2 \\
     \hline  
     stride=1 & \bf 61.01  & \bf 69.14  & \bf 69.41 \\
     stride=2 & 59.56  & 60.26  & 63.46 \\
     stride=3 & 42.35  & 54.32  & 57.36 \\
     stride=4 & 46.07  & 55.12  & 54.82 \\
     \hline
  \end{tabular} 
  \vspace{2mm}
  \caption{Influence of patch overlap. We control the patch overlap by using different strides at the prediction layer of the patch discriminator.} \label{tbl:overlap}
  \vspace{-2mm}
\end{table}

\subsubsection{Patch Overlap} 
Given a fixed patch size, 
the overlap between patches
plays an important role in patch learning.
The overlap is controlled by 
interval sampling in the patch discriminator. 
Note that the discriminator produces a prediction map to predict each small region on the original image, which means that distant predictions should share less information. We add a prediction stride to the final discrimination to control the patch overlap. Table \ref{tbl:overlap} shows the student accuracy obtained
with different patch overlaps, where a larger stride corresponds to a smaller overlap. The results show that increasing stride does not benefit the students' accuracy. Note that we use the patch GAN architecture for patch learning, which contains internal stride operations within the discriminator. These stride operations already provide an appropriate overlap for patch learning. Besides, a larger stride also means fewer training samples, which may be harmful to the GAN training.

\subsubsection{DRO Regularization}
\begin{figure}[t]
  \begin{center}
     \includegraphics[width=\linewidth]{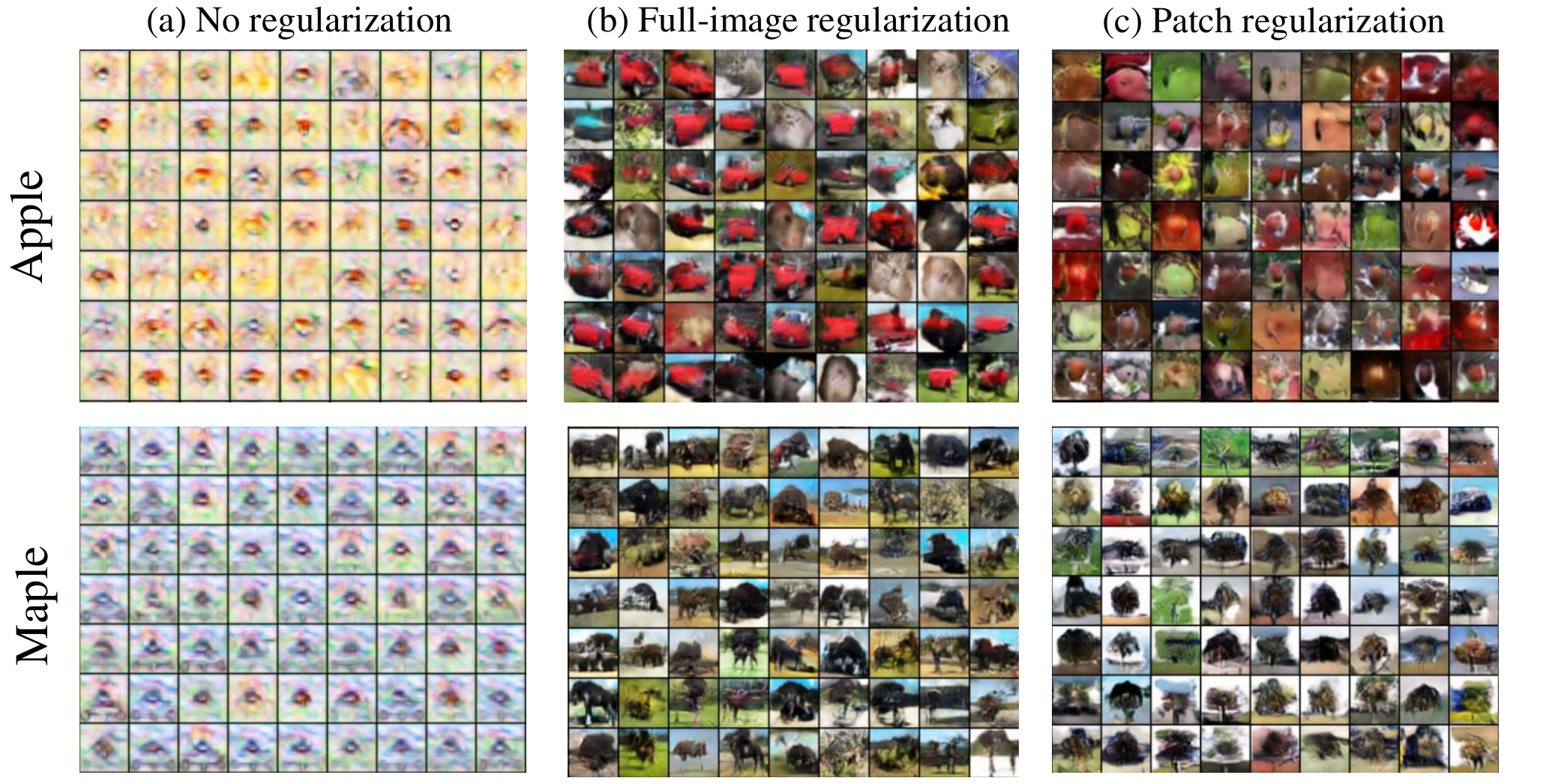}
  \end{center}\vspace{-3mm}
     \caption{Synthetic images from the generator: (a)  without regularization, (b) with full image regularization, and (c) with patch regularization.} \label{fig:reg}
\end{figure}

In MosaicKD, the search space is regularized by $\mathcal{L}_{local}$ and $\mathcal{L}_{align}$, which enforces the generated samples to be locally authentic and globally legitimate. We take a further study on the above regularization to show their significance for MosaicKD. As 
illustrated in~\ref{fig:reg}, we visualize the
generated samples with different regularizations.  
In Figure~\ref{fig:reg}(a), 
no regularization is applied on the generator, 
and we naively maximize the teacher's confidence, 
which will lead to some inferior 
samples~\cite{goodfellow2014explaining}. In Figure~\ref{fig:reg}(b), the discriminator makes decisions on full images, and, to some extent, the generator will be trapped by the class semantic of OOD data, i.e., synthesizing a car-like apple or a horse-like maple. Figure(c) showcases the synthetic samples of MosaicKD, which reveals the correct semantic of task-related classes. 

\subsubsection{ImageNet Results}
Table \ref{tbl:imagenet} provides the student's accuracy on $32\times 32$ ImageNet dataset with 1000 categories. We use Places365~\cite{zhou2017places} as the OOD data and resize all samples to $32\times 32$ for training. Results show that our approach is indeed beneficial for the OOD-KD task.

\begin{table*}[h]
  \centering
  \begin{tabular}{l c c c}
      \toprule
      \bf \multirow{2}{*}{Method} & \bf \multirow{2}{*}{Data} & \bf resnet-56  & \bf resnet-56 \\
                                  &                           & \bf resnet-20  & \bf mobilenetv2 \\                  
    \hline  
      Teacher & \multirow{3}{*}{\shortstack{ImageNet\\(Original Data)}} & 41.28 & 41.28 \\ % lr 0.1
      Student &                       & 32.20 & 32.48  \\
      KD~\cite{hinton2015distilling}   &  & 32.18 & 32.55 \\
      \hline
      KD~\cite{hinton2015distilling} &  \multirow{4}{*}{\shortstack{Places365\\(OOD Data)}} & 21.76  & 10.25 \\
      Balanced~\cite{nayak2021effectiveness} &  & 21.09 & 11.34  \\
      FitNet~\cite{romero2014fitnets} &  &  21.45 & 13.12 \\
      Ours & & \bf 26.51  & \bf 20.46  \\
      \hline
  \end{tabular} 
  \vspace{2mm}
  \caption{Test accuracy (\%) of student networks on ImageNet. We use the full places365 dataset as transfer set for OOD-KD.}
  \vspace{-2mm}
  \label{tbl:imagenet}
\end{table*} 

\end{document}